\documentclass[10pt]{article} 

\usepackage[preprint]{rlj}           

\usepackage{amssymb}            
\usepackage{mathtools}          
\usepackage{mathrsfs}           
\usepackage{graphicx}           
\usepackage{subcaption}         
\usepackage[space]{grffile}     
\usepackage{url}                
\usepackage{lipsum}             
\usepackage{booktabs}
\usepackage[linesnumbered,ruled,vlined]{algorithm2e}
\usepackage{caption}
\usepackage{float}
\usepackage{listings}
\definecolor{codegreen}{rgb}{0,0.6,0}
\definecolor{codegray}{rgb}{0.5,0.5,0.5}
\definecolor{codepurple}{rgb}{0.58,0,0.82}
\definecolor{backcolour}{rgb}{0.95,0.95,0.92}

\lstdefinestyle{mystyle}{
    backgroundcolor=\color{backcolour},   
    commentstyle=\color{codegreen},
    keywordstyle=\color{magenta},
    numberstyle=\tiny\color{codegray},
    stringstyle=\color{codepurple},
    basicstyle=\ttfamily\footnotesize,
    breakatwhitespace=false,         
    breaklines=false,                 
    captionpos=b,                    
    keepspaces=true,                 
    numbers=none,                    
    numbersep=3pt,                  
    showspaces=false,                
    showstringspaces=false,
    showtabs=false,                  
    tabsize=2
}

\DeclareMathOperator*{\argmax}{arg\,max}

 \title{Evaluating Interpretable Reinforcement Learning by Distilling Policies into Programs}

\setrunningtitle{Evaluating Interpretable RL Policies}

\author{Hector Kohler\textsuperscript{1,2,3,$\dagger$}, Quentin Delfosse\textsuperscript{4}, Waris Radji\textsuperscript{2,1,3},
Riad Akrour\textsuperscript{2,1,3},
Philippe Preux\textsuperscript{1,2,3},}

\emails{hector.kohler@inria.fr, \ quentin.delfosse@cs.tu-darmstadt.de, \ 
\{waris.radji,riad.akrour,philippe.preux\}@inria.fr}

\affiliations{
$^{1}$\textbf{Universit\'e de Lille, France}\\
$^{2}$\textbf{Inria, France}\\
$^{3}$\textbf{UMR 9198-CRIStAL, CNRS, Centrale Lille, France}\\
$^{4}$\textbf{AI \& ML Lab, Computer Science Department, TU Darmstadt, Germany}\\
$\dagger$ Corresponding author}

\contribution{
    We provide a sound methodology to evaluate the interpretability of policies for Markov decision processes without human studies.
    }
    {
    There is no methodology to study interpretability without humans that both, supports results that interpretability depends on the policy class and on the number of policy parameters \citep{study-0,study-1,study-2,study-3,study-4}, and, claims that policy interpretability depends on the chosen metrics \citep{mythos}. Hence providing such a methodology is an open problem in interpretable reinforcement learning \citep{glanois-survey}.
    }

\contribution{
    Using our methodology, we show that, across the tested tasks, and across tested interpretability metrics, there is no policy class that is better in terms of cumulative reward, or better in terms of interpretability, or better in trading off reward and interpretability. 
    }
    {
    Human studies have shown that for particular users and particular tasks, trees are considered more interpretable than policies involving equations \citep{study-1,study-2,study-3}. However, \citet{mythos} argues that this claim depends on the choice of interpretability definition. Since there is no definite interpretability ranking of policy classes, having a methodology to compare the interpretability-performance trade-offs of policies of different classes is crucial.
    }

\contribution{
    We show that to properly evaluate policy interpretability it is necessary to ``unfold'' policies into a similar language such as Python to standardize the execution of policy inference.  
    }
    {
    Previous works have attempted to compare policies interpretability without grounding them into a common language leading to results that depend on hardware and software choices \citep{viper,insight}.
    }

\contribution{
    We open source the interpretable policies that we use to validate our methodology: \url{https://github.com/KohlerHECTOR/interpretable-rl-zoo}.
    }
    {
    Similarly to open source baselines for reinforcement learning \citep{sb3,zoo}, we hope to advance research in interpretable reinforcement learning.
    }


\summary{There exist applications of reinforcement learning like medicine where policies need to be ``interpretable'' by humans. User studies have shown that some policy classes might be more interpretable than others. However, it is costly to conduct human studies of policy interpretability. Furthermore, there is no clear definition of policy interpretabiliy, i.e., no clear metrics for interpretability and thus claims depend on the chosen definition. We tackle the problem of empirically evaluating policies interpretability without humans. Despite this lack of clear definition, researchers agree on the notions of ``\textit{simulatability}'': policy interpretability should relate to how humans understand policy actions given states. To advance research in interpretable reinforcement learning, we contribute a new methodology to evaluate policy interpretability. We use our methodology to conduct a large-scale empirical evaluation of policy interpretability and study how our empirical conclusions relate to previous work.}

\begin{document}

\maketitle  
\begin{abstract}
There exist applications of reinforcement learning like medicine where policies need to be ``interpretable'' by humans. User studies have shown that some policy classes might be more interpretable than others. However, it is costly to conduct human studies of policy interpretability. Furthermore, there is no clear definition of policy interpretabiliy, i.e., no clear metrics for interpretability and thus claims depend on the chosen definition. We tackle the problem of empirically evaluating policies interpretability without humans. Despite this lack of clear definition, researchers agree on the notions of ``\textit{simulatability}'': policy interpretability should relate to how humans understand policy actions given states. To advance research in interpretable reinforcement learning, we contribute a new methodology to evaluate policy interpretability. This new methodology relies on proxies for \textit{simulatability} that we use to conduct a large-scale empirical evaluation of policy interpretability. We use imitation learning to compute baseline policies by distilling expert neural networks into small programs. We then show that using our methodology to evaluate the baselines interpretability leads to similar conclusions as user studies. We show that increasing interpretability does not necessarily reduce performances and can sometimes increase them. We also show that there is no policy class that better trades off interpretability and performance across tasks making it necessary for researcher to have methodologies for comparing policies interpretability. 
\end{abstract}

\section{Introduction}
There is increasing research in developing reinforcement learning algorithms that return ``interpretable" policies such as trees, programs, first-order logic, or linear maps \citep{viper,PIRL,empirical-evidence,nudge,milani-survey,glanois-survey}. Indeed, interpretability has been useful for different applications: policy verification \citep{viper}, mis-alignment detection \citep{scobots,sympol} and features importance analysis \citep{fi-rl,fi-rl2,fi-rl3}.

User studies have established the common beliefs that decision trees are more ``interpretable" than linear maps, oblique trees, and multi-layer perceptrons (MLPs)~\citep{study-0,study-1,study-2,study-3}. Furthermore, for a fixed class of models, humans give different values of interpretability to models with different numbers of parameters~\citep{study-4}. However, survey works argue that every belief about interpretability needs to be verified with user studies and that interpretability evaluations are grounded to a specific set of users, to a specific application, and to a specific definition of interpretability \citep{rigorous,mythos}. For example, \citet{mythos} claims that depending on the notion of \textit{simulatability} studied, MLPs can be more interpretable than trees, since deep trees can be harder for a human to read than compact MLPs. Hence, even with access to users it would be difficult to research interpretability. More realistically, since the cost of user studies is high (time, variety of subjects required, ethics, etc.), designing proxies for interpretability in machine learning has become an important open problem in both supervised~\citep{rigorous} and reinforcement learning~\citep{glanois-survey}.

In this work, we propose a methodology to evaluate the interpretability of reinforcement learning policies without human evaluators, by measuring inference time and memory consumption. We show that those measures constitute adequate proxies for the notions of ``\textit{simulatability}'' described in~\citet{mythos}, which relates the interpretability of policy to humans ability to understand the inference of actions given states. In addition to the contributions summarized next, we open source some of the interpretable baselines to be be used for future interpretability research and teaching\footnote{\url{https://github.com/KohlerHECTOR/interpretable-rl-zoo}}.


In Section \ref{sec:res-imit}, we distill deep neural network experts into more compact programs that can be used as baselines for interpretability evaluation. In Section \ref{sec:ablation-metric}, we validate the proposed methodology and show that it is necessary to ground policies into a common language to obtain meaningful interpretability measures. In Section \ref{sec:res-trade-offs},
we show how our methodology can be used to tackle interpretable reinforcement learning problems : I) Is it possible to obtain a fully interpretable policy for MDPs with large state-action spaces? II) Do policies have to trade off rewards for interpretability? III) Can interpretable policies be formally verified efficiently?

\section{Related Work}
Evaluating policy interpretability without humans should be possible using the common beliefs stated above. However, many related works do not do so correctly. A first category of research works do not even attempt to evaluate interpretability while claiming it. For example, ~\citet{theory1} show that there always exists a small oblique decision tree \citep{oblique} that can optimally solve continuous-states mazes. 
Because the policy they consider is a tree and not an MLP, they claim that there is not necessarily a trade-off between interpretablity and performance in mazes. But how can they ensure that such trees are interpretable? A second category of work compares policies from a fixed class using the belief that less parameters means more interpretability. 
\citet{viper,IBMDP,sympol} learn tree policies and use the number of nodes in their policies as a metric for interpretability. However, not all nodes of decision trees are traversed during inference, hence a human's ability to understand policy predictions can not solely depend on the number of parameters of policies \citep{mythos}.
The last category of work attempts to evaluate the interpretability of different policy classes using proxies. 
We argue that they do not obtain meaningful conclusions because they do not account for policy implementation details nor software and hardware differences. \citet{viper} claim that trees are more interpretable than MLPs because it is faster to do formal verification of the former. However, it is hard to draw conclusions from this comparison as the software for formal verification of trees \citep{z3} is very different from the one to verify MLPs \citep{maraboupy}. Similarly, \citet{insight} compares the interpretability of MLPs and of first-order logic rules \citep{nudge}. They use inference speed as a proxy without grounding both policies in a common language. This evaluation is biased, as one policy can make use of speed optimization such as array operations or GPU computations, while the other could be run completely sequentially. 
\citet{zahra} propose to measure the interpretability of policies implemented similarly as programs using the perplexity of a large language model tasked with explaining the policies. 
However, querying large language models is not efficient when doing large-scale experiments. 
In the next section, we propose a new methodology to properly evaluate the interpretability of policies from different classes.

\section{Methodology}\label{sec:methodology}
In this section, we explain our methodology for the evaluation of interpretability.
Our approach consists of three main steps: (1) obtaining deep neural network policies trained with reinforcement learning that obtain high cumulative rewards, (2) distilling those policies into less complex ones to use as baselines (3) after parsing baselines from different classes into a common comparable language, we evaluate the interpretability of the policies using proxy metrics for \textit{simulatability}.

\paragraph{Deep Neural Network Policies}
In reinforcement learning, an agent learns how to behave in an environment to maximize a cumulative reward signal~\citep{sutton}. 
The environment is defined by a Markov decision process (MDP) $M = ( \mathcal{S}, \mathcal{A}, T, R)$, where $\mathcal{S}$ is the state space, $\mathcal{A}$ is the action space, $T : \mathcal{S} \times \mathcal{A} \times \mathcal{S} \rightarrow [0, 1]$ is the state-transition function, $R : \mathcal{S} \times \mathcal{A} \times \mathcal{S} \rightarrow \mathbb{R}$ is the reward function.
At each time step $t$, an agent observes the current state $s_t$ and chooses an action $a_t$ according to its policy $\pi$.
The agent executes $a_t$, receives reward $r_t$, and observes the next state $s_{t+1}'$.
The goal is to find an optimal policy $\pi^*$ that maximizes the expected discounted future return: $\pi^* = \argmax_{\pi} Q^\pi (s, a) =\argmax \mathbb{E}[r+\gamma Q^\pi(s', a)]$, with $\gamma$ a discount factor in $[0,1)$. For large or continuous state spaces like the MDPs we consider in this work, MLPs are used to represent $Q^\pi$ or $\pi$.
While these MLPs can be trained efficiently to obtain high cumulative rewards~\citep{ppo,dqn}, they are too complex for interpretability considerations.

\paragraph{Distilling into Interpretable Policies}
To obtain interpretable policies, we distill the complex neural networks into simpler models using imitation learning, as described in Algorithm \ref{alg:distill}. This approach transforms the reinforcement learning task into a sequence of supervised learning problems.

\begin{algorithm}
\caption{Imitate Expert~\citep{behavior-cloning,dagger,viper}}\label{alg:distill}
\KwIn{Expert policy $\pi^*$, MDP $M$, policy class $\Pi$, number of iterations $N$, total samples $S$, importance sampling flag $I$}
\KwOut{Fitted student policy $\hat{\pi}_i$}

Initialize dataset $\mathcal{D} \gets \emptyset$\;
Initialize $\hat{\pi}_1$ arbitrarily from $\Pi$\;

\For{$i \gets 1$ \KwTo $N$}{
    \lIf{i = 1}{
    $\pi_i \gets \pi^*$\
    }
      \lElse{
        $\pi_i \gets \hat{\pi}_i$
      } \label{alg:expert-or-student}
    Sample $S/N$ transitions from $M$ using $\pi_i$\;
    \lIf{$I$ is True}{
    $w(s) \gets V^{\pi*}(s) - \operatorname{min}_aQ^{\pi*}(s, a)$
    }
      \lElse{
        $w(s) \gets 1$
      }
    Collect dataset $\mathcal{D}_i \gets \{ (s, \pi^*(s), w(s)) \}$ of states visited by $\pi_i$ and expert actions\;\label{alg:actions-expert}
    
    $\mathcal{D} \gets \mathcal{D} \cup \mathcal{D}_i$\;
    
   Fit classifier/regressor $\hat{\pi}_{i+1}$ on $\mathcal{D}$\;\label{alg:fit}
}

\Return $\hat{\pi}_N$\;
\end{algorithm}
Algorithm \ref{alg:distill} inputs an environment, that simulates taking steps in an MDP, an expert policy to imitate, also called a teacher, and an (interpretable) policy class to fit, also called student. The hyperparameters of Algorithm \ref{alg:distill} are: the number of times we fit a student policy, the total number of samples to be collected, and whether or not to use importance sampling. At each iteration of Algorithm \ref{alg:distill} the student policy is fitted to a dataset of states collected with the expert at iteration 1 or with the previously fitted student (see Line \ref{alg:expert-or-student}). The actions are always given by the expert (see Line \ref{alg:actions-expert}). When using importance sampling, the states are further re-weighted by the worst state-action value possible in the given state. When the number of iteration is 1, Algorithm \ref{alg:distill} is behavior cloning~\citep{behavior-cloning}. When we use importance sampling, Algorithm \ref{alg:distill} is $Q$-DAgger~\citep{viper}. In other cases, Algorithm \ref{alg:distill} is DAgger~\citep{dagger}.

\paragraph{Measuring Policy Interpretability}
After obtaining interpretable policy baselines using Algorithm \ref{alg:distill}, we use two metrics to evaluate policy interpretability without requiring human intervention. Those metrics are proxies for the notion of \textit{simulatability} from~\citet{mythos} that gives insights on how a human being would read a policy to understand how actions are inferred. In particular, \textit{simulatability} admits two sub-definitions. The first one is a measure of how difficult it is for a human to reproduce the computations of the policy to infer actions given states. The second one measures how difficult it is for a human to read through the entire policy. \citet{mythos} argues that this nuance is key when measuring interpretability because a tree is not read entirely to compute a single action and because there is no consensus on what is easier for a human to read between an MLP and a tree. 

\textit{1. Policy Inference Time:} to measure how a human would compute the action of a policy given a state at each environment step, we measure policy step inference time in seconds.

\textit{2. Policy Size:} to measure how easily a human can read the entire policy, we measure its size in bytes. While this correlates with inference time for MLPs and linear models, tree-based policies may have large sizes but quick inference because they do not traverse all decision paths at each step.

As these measurements depend on many technical details (programming language, the compiler if any, the operating system, versions of libraries, the hardware it is executed on, etc), to ensure fair comparisons, we translate all student policies into a simple representation that mimics how a human being "reads" a policy. We call this process of standardizing policies language ``unfolding''.
This unfolding allows for meaningful comparisons across different policy types and is illustrated in code snippets \ref{lst:generic-linear} and \ref{lst:unfolded-linear}. Other works have distilled neural networks into programs \citep{PIRL} or even directly learn programmatic policies \citep{pirl2} from scratch. However, those works directly consider programs as a policy class and could compare a generic program (not unfolded, with, e.g., while loops or array operations) to, e.g, a decision tree \citep{leap}. We will discuss later on the limitations of unfolding policies in the overall methodology.

\begin{figure}
\centering
\begin{minipage}{0.42\textwidth}
\begin{tcolorbox}
\begin{lstlisting}[language=Python]
import gymnasium as gym
from pickle import load

policy = \
load("policy_mountaincar.pkl")

env = gym.make("MountainCar")
s, _ = env.reset()
done = False
while not done:
    action = policy.predict(s)
    s, r, term, truncated, \ 
    infos = env.step(action)
    done = term or truncated
\end{lstlisting}
\end{tcolorbox}
\caption{Generic linear policy with hidden operations interacting with an environment.}\label{lst:generic-linear}
\end{minipage}
\hfill
\begin{minipage}{0.54\textwidth}
\begin{tcolorbox}
\begin{lstlisting}[language=Python]
import gymnasium as gym

env = gym.make("MountainCar")
s, _ = env.reset()
done = False
while not done:
    y0 = 0.969*s[0] -30.830*s[1] +0.575
    y1 = -0.205*s[0] + 22.592*s[1] -0.63
    y2 = -0.763*s[0] + 8.237*s[1] +0.054
    max_val = y0
    action = 0
    if y1 > max_val:
        max_val = y1
        action = 1
    if y2 > max_val:
        action = 2
    s, r, terminated, truncated, \
    infos = env.step(action)
    done = terminated or truncated
\end{lstlisting}
\end{tcolorbox}
\caption{Unfolded linear policy interacting with an environment.}\label{lst:unfolded-linear}
\end{minipage}
\end{figure}

\section{Results}

In this section, we follow the methodology of Section \ref{sec:methodology} and evaluate the interpretability of different policy classes using proxies of \textit{simulatability}. All the experiments presented next run on a dedicated cluster of Intel Xeon Gold 6130 (Skylake-SP), 2.10GHz, 2 CPUs/node, 16 cores/CPU with a timeout of 4 hours per experiment. Codes to reproduce our results are given in the supplementary material. In the future, we will open source a python library with all the tools of our methodology.

\subsection{Baselines}
Using Algorithm \ref{alg:distill}, we distill deep neural network expert policies into less complex policy classes.

\paragraph{Policy classes}

 We consider four policy classes for our baselines. We choose those policy classes because there exist efficient algorithms to fit them with supervised data which is a required step of imitation learning in Line \ref{alg:fit}. We consider linear policies that have been shown to be able to solve Mujoco tasks~\citep{empirical-evidence}. We fit linear policies to expert policies using simple linear (logistic) regressions with scikit-learn~\citep{scikit-learn} default implementation. We also consider decision trees~\citep{cart} and oblique decision trees~\citep{oblique}. (Oblique) Decision trees are often considered the most interpretable model class in machine learning~\citep{mythos} and reinforcement learning~\citep{viper,IBMDP,glanois-survey,milani-survey}. We train trees using the default CART~\citep{cart} implementation of scikit-learn with varying numbers of parameters (number of nodes in the tree). We also consider MLPs with ReLU activations~\citep{relunet} with varying number of parameters (total number of weights). This class of policy is often considered the least interpretable and is often used in deep reinforcement learning~\citep{deep-rl-relu1,deep-rl-relu2,deep-rl-relu3}. We train ReLU MLPs using the default scikit-learn implementation of Adam optimization~\citep{adam} with 500 iterations. The 15 baseline policy classes that we consider are summarized in Appendix \ref{tab:policy-classes}. 
\paragraph{Neural network experts}
We do not train new deep reinforcement learning agents \citep{dqn,ppo,deep-rl-relu1} but rather re-use ones available at the stables-baselines3 zoo \citep{zoo}. Depending on the environments described next, we choose neural network policies from different deep reinforcement learning agents. Some may argue that during the imitation learning, ReLU MLPs baselines may obtain better performance because they are often from the same class as the expert they imitate unlike trees. But this is not of our concern as we do not benchmark the imitation learning algorithms. Furthermore, it is important to note that not all experts are compatible with all the variants of imitation learning Algorithm \ref{alg:distill}. Indeed, SAC experts \citep{deep-rl-relu1} are not compatible with $Q$-DAgger \citep{viper} because it only works for continuous actions; and PPO experts, despite working with discrete actions do not compute a $Q$-function necessary for the re-weighting in $Q$-DAgger.

\paragraph{Environments}
We consider common environments in reinforcement learning research. We consider the classic control tasks from gymnasium \citep{gymnasium}, MuJoCo robots from \citet{mujoco}, and Atari games from \citet{atari}. For Atari games, since the state space is frame pixels, we use the object-centric version of the games from \citet{ocatari}. In Appendix \ref{tab:envs} we give the list of environments we consider in our experiments with their state-action spaces as well as a cumulative reward threshold past which an environment is consider ``solved''.

\subsection{Ablation study of imitation learning}\label{sec:res-imit}

In this section, we present the results of the expert distillation into smaller policies. For each environment, we fit all the policy classes. To do so, we run different instantiations of Algorithm \ref{alg:distill} multiple times with different total sample sizes. For each environment and each imitation learning variant, we summarize the number of times we fit all the baselines to an expert and which expert we use. The number of runs and imitation algorithm variants of Algorithm \ref{alg:distill} are summarized in Appendix \ref{tab:repet-distill}. After running the imitation learnings, we obtain roughly 40000 baseline policies (35000 for classic control, 5000 thousands for MuJoCo and 400 for OCAtari). A dataset with all the baselines measurements is given in the supplementary material.

\begin{figure}[ht]
\centering
\begin{subfigure}{.33\textwidth}
  \centering
  \includegraphics[width=\linewidth]{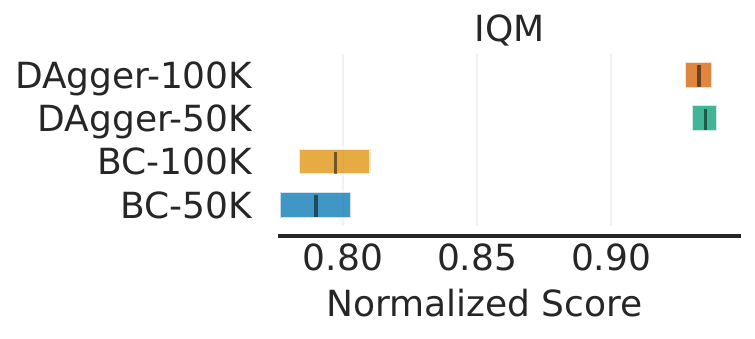}
  \caption{Classic control, PPO expert}
  \label{fig:ppo_classic}
\end{subfigure}%
\begin{subfigure}{.33\textwidth}
  \centering
  \includegraphics[width=\linewidth]{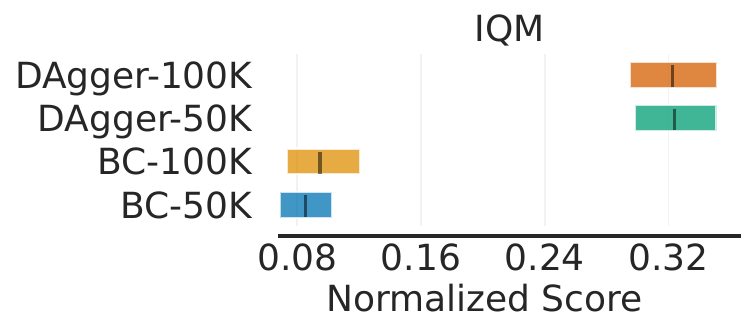}
  \caption{MuJoCo, SAC expert}
  \label{fig:sac_mujoco}
\end{subfigure}
\begin{subfigure}{.33\textwidth}
  \centering
  \includegraphics[width=\linewidth]{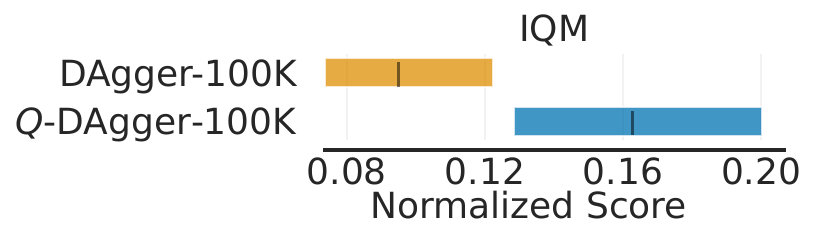}
  \caption{OCAtari, DQN expert}
  \label{fig:dqn_atari}
\end{subfigure}%
\caption{Performance of imitation learning variants of Algorithm \ref{alg:distill} on different environments. We plot the 95\% stratified bootstrapped confidence intervals around the IQMs.}
\label{fig:performance_comparison}
\end{figure}

\paragraph{What is the best imitation algorithm?}
Even though the focus of our work is to evaluate trained policies, we still provide some insights on the best way to obtain interpretable policies from experts. Using the reinforcement learning evaluation library rliable~\citep{rliable}, we plot on Figure \ref{fig:performance_comparison} the interquartile means (IQM, an estimator of the mean robust to outliers) of the baseline policies cumulative rewards averaged over 100 episodes. For each imitation algorithm variant, we aggregate cumulative rewards over environments and policy classes. We normalize the baselines cumulative rewards between expert and random agent cumulative rewards.

The key observation is that for tested environments (Figures \ref{fig:ppo_classic},\ref{fig:sac_mujoco}), Behavior Cloning is not an efficient way to train baseline policies compared to DAgger. This is probably because Behavior Cloning trains a student policy to match the expert's actions on states visited by the expert while DAgger trains a student to take the expert's actions on the states visited by the student~\citep{dagger}. An other observation is that the best performing imitation algorithms for MuJoCo (DAgger, Figure \ref{fig:sac_mujoco}) and OCAtari ($Q$-Dagger, Figure \ref{fig:dqn_atari}) obtain baselines that in average cannot match well the performances of the experts. However baseline policies almost always match the expert on simple tasks like classic control (Figure \ref{fig:ppo_classic}).

\paragraph{What is the best policy class in terms of reward?}

\begin{figure}[ht]
    \centering
    \includegraphics[trim={0 0 0 0.2cm},clip,width=1\linewidth]{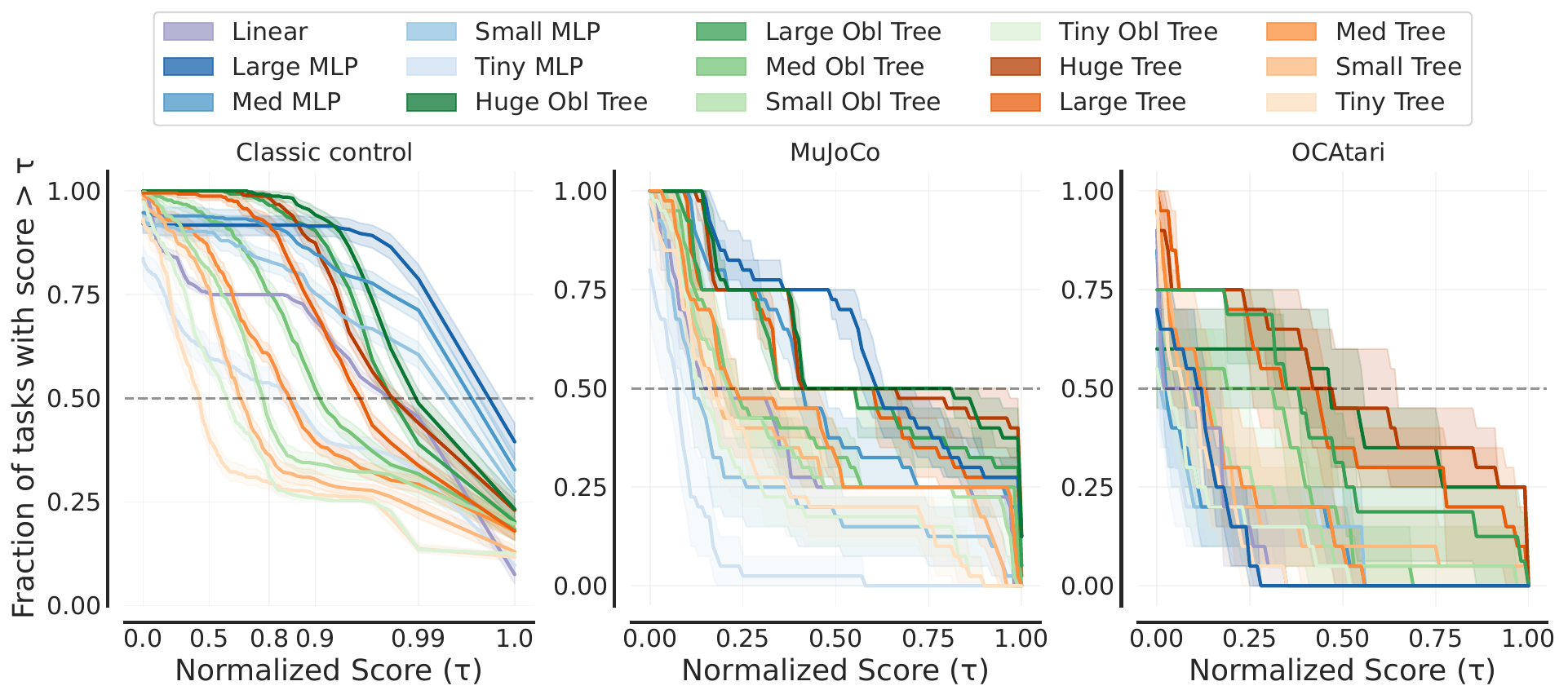}
    \caption{Performance profiles of different policy classes on different environments.}
    \label{fig:perf-combined}
\end{figure}

We also wonder if there is a policy class that matches expert performances more often than others across environments. For that we plot performance profiles of the different policy classes obtained with a fixed expert and fixed imitation learning algorithm. In particular, for each environments group we use the baseline policies obtained from the best performing imitation learning algorithm from Figure \ref{fig:performance_comparison}. From Figure \ref{fig:perf-combined} we see that on classic control environments, MLPs tend to perform better than other classes while on OCAtari games, trees tend to perform better than other classes. Now we move on to the interpretability evaluation with proxies of \textit{simulatability}.

\subsection{Validation of the proposed methodology}\label{sec:ablation-metric}

\begin{figure}
    \centering
    \includegraphics[width=1\linewidth]{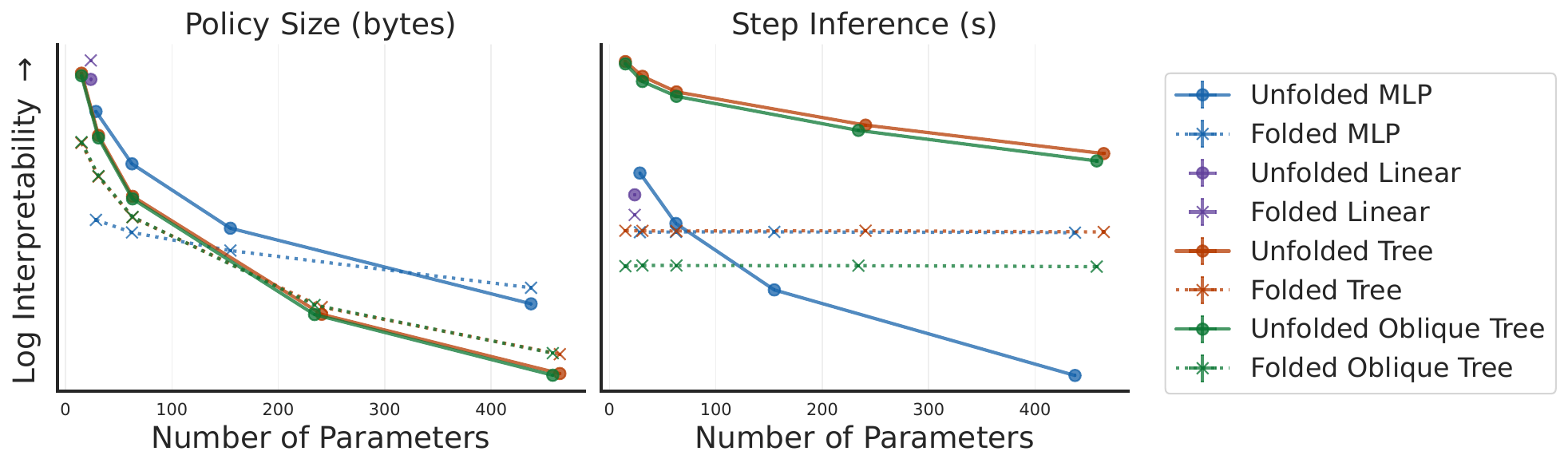}
    \caption{\textit{simulatability} proxies on classic control environments. We plot 95\% stratified bootstrapped confidence intervals around means in both axes. In each sub-plot, interpreatbility is measured with the proxy corresponding to the sub-title.}
    \label{fig:abl-proxies}
\end{figure}

In this section, we aim to validate the proposed methodology from Section \ref{sec:methodology}. We compute the step inference times, as well as the policy size for both the folded and unfolded variant of each policy obtained for classic control environments with DAgger-100K. To unfold policies, we convert them into Python programs formatted with PEP 8. We ensure that all policies operations are performed sequentially and compute the metrics for each policy on 100 episodes using the same CPUs.

\paragraph{Is it necessary to unfold policies to compute interpretability metrics?} We see on Figure \ref{fig:abl-proxies} that folded policies of the same class almost always give similar interpretability values (dotted lines) despite having very different number of parameters. Hence, measuring folded policies interpretability would contradict established results from user studies such as, e.g., trees of different sizes have different levels of interpretability~\citep{study-4}. 

\paragraph{Is there a best policy class in terms of interpretability?}
User studies from~\citet{study-1,study-2,study-3} show that decision trees are easier to understand than models involving mathematical equations like oblique trees, linear maps, and MLPs. However,~\citet{mythos} states that for a human wanting to have a global idea of the inference of a policy, a compact MLP can be more interpretable than a very deep decision tree. In Figure \ref{fig:abl-proxies}, we show that our proposed proxies help us capture those nuances: policy interpretability does not only depend on the policy class but also on the metric choice. Indeed, when we measure interpretability with inference times, we do observe that trees are more interpretable than MLPs. However, when measuring interpretability with policy size, we observe that MLPs can be more interpretable than trees for similar number of parameters. Because there seem to not be a more interpretable policy class across proxy metrics, we will keep studying both metrics at the same time.

\subsection{Interpretability-performance trade-offs}\label{sec:res-trade-offs}

Now that we trained baseline policies and validated the proposed methodology, we use the latter to tackle open problems in interpretable reinforcement learning. For each environment, we fix the imitation learning algorithm and save the best baseline policy of each class in terms of episodic rewards after unfolding them. Each single Python policy is then \textbf{run again on the same dedicated CPU} for 100 new environment episodes (similarly to choosing a classifier with validation score and reporting the test score in the context of supervised learning).
\begin{figure}
    \centering
    \includegraphics[trim={1.4cm 0 0 0},clip,width=1\textwidth]{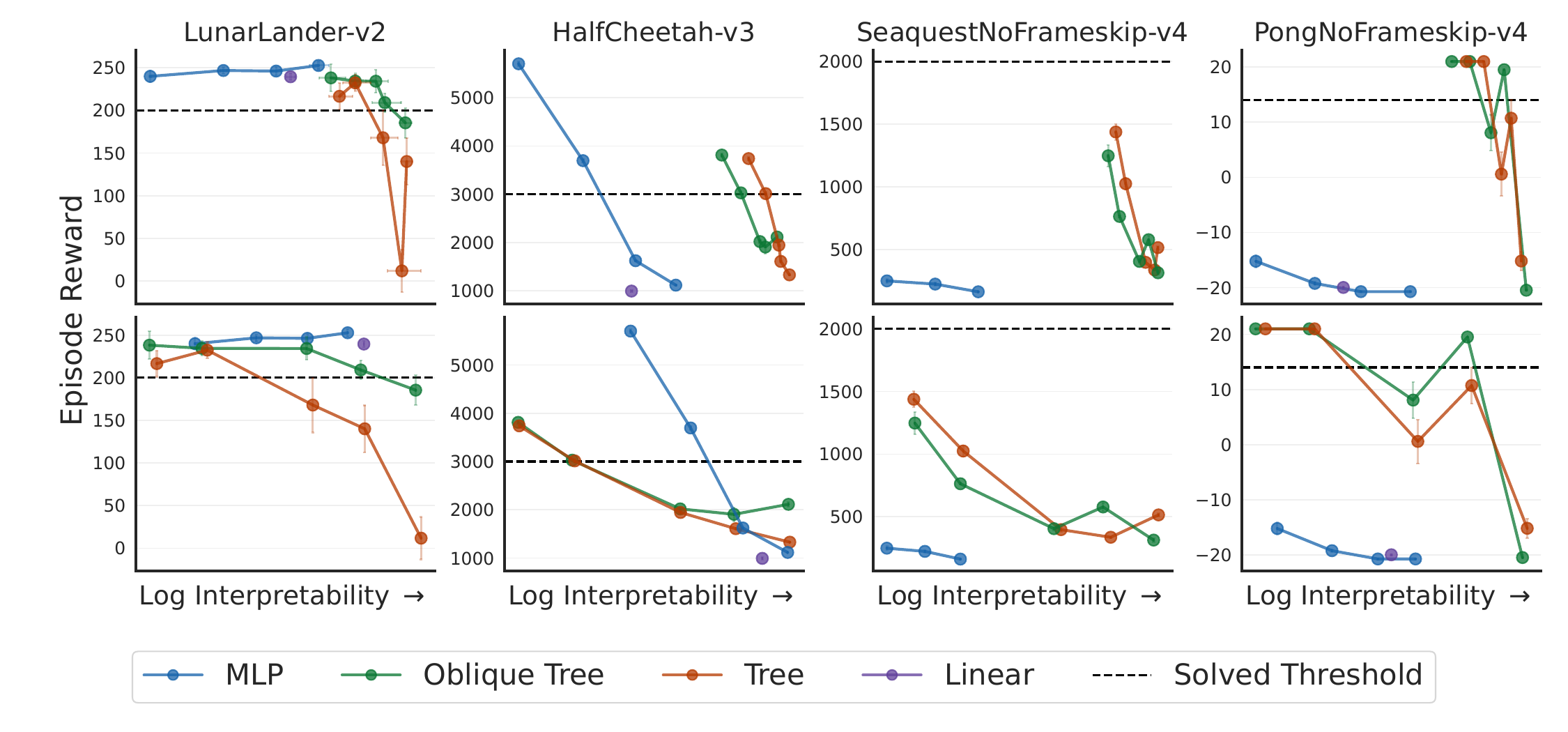}
    \caption{Interpretabality-Performance trade-offs. Top row, interpretability is measured with step inference times. Bottom row, the interpretability is measured with policy size. We plot 95\% bootstrapped confidence intervals around means on both axes.}
    \label{fig:trade-off-summary}
\end{figure}

\paragraph{Is it possible to compute interpretable policies for high-dimensional environments?} \citet{glanois-survey} claim that computing an interpretable policy for high dimensional MDPs is difficult since it is similar to program synthesis which is known to be NP-hard \citep{program-synth}. Using our measures of interpretability, we can corroborate this claim. On Figure \ref{fig:trade-off-summary}, we can indeed observe that some relatively interpretable policies can solve Pong (20 state dimensions) or HalfCheetah (17 state dimensions) while for very high-dimensional environments like Seaquest (180 state dimensions), no baseline can solve the game. Performance-interpretability trade-offs are given for all environments in Figures \ref{fig:trade-off} and \ref{fig:trade-off-size}.

\paragraph{For what environment are there good interpretable policies?}
We fitted a random forest regressor \citep{random} to predict the interpretability values of our baseline policies using environment attributes. In Table \ref{tab:combined_importance} we report the importance of each environment attribute when it comes to accurately predicting interpretability scores. We show that as hinted previously, the states dimensions of the environment is determining to predict the interpretability of good policies. Unsurprisingly, expert attributes also influence interpretability: for the environments where there is a positive large gap between expert and threshold rewards, the task could be considered easy and vice-versa.

\begin{table}
\centering
\small
\begin{tabular}{lcc}
\toprule
Environment Attributes & Importance for Step inference & Importance for Policy size \\
\midrule
States dimension & \textbf{80.87} & \textbf{35.52} \\
Expert episodes lengths & 11.39 & 9.28 \\
Episode reward of random & 2.26 & 4.75 \\
Expert episode reward & 1.51 & 16.80 \\
Episode reward to solve & 1.41 & 14.26 \\
Actions dimension & 1.41 & 2.02 \\
Expert reward - Solve reward & 1.15 & 17.37 \\
\bottomrule
\end{tabular}
\caption{Environment attributes importance to predict interpretability using either of our metrics.}
\label{tab:combined_importance}
\end{table}

\paragraph{How does interpretability influence performance?}
\citet{empirical-evidence,theory1} show the existence of linear and tree policies respectively that solve MuJoCo and continuous maze environments respectively; essentially showing that there exist environments for which policies more interpretable than deep neural networks can still compete performance-wise. Our evaluation indeed shows the existence of such environments. On Figure \ref{fig:trade-off-summary} we observe that on, e.g., 
LunarLander, increasing policy interpretability up to a certain point does not decrease reward. Actually, we can observe that for Pong a minimum level of interpretability is required to solve the game. Indeed, as stated in \citep{study-0}, optimizing interpretability can also be seen as regularizing the policy which can increase generalization capabilities. 
The key observation is that the policy class achieving the best interpretability-performance trade-off depends on the problem. Indeed, independent of the interpretability proxy metric, we see on Figure \ref{fig:trade-off-summary} that for LunarLander it is an MLP that achieves the best trade-off while for Pong it is a tree. Next, we compare our proxies for interpretability with another one; the verification time of policies used in \citet{viper,lens-complexity}.

\subsection{Verifying interpretable policies}
\citet{lens-complexity} states that the cost of formally verifying properties of MLPs scales exponentially with the number of the parameters. Hence, they propose to measure interpretability of a policy as the computations required to verify properties of actions given state subspaces, what they call local explainability queries \citep{query}. Before \citet{lens-complexity}, \citet{viper} also compared the time to formally verified properties of trees to the time to verify properties of MLPs to evaluate interpretability. In practice, this amounts to passing states and actions bounds and solving the SAT problem of finding a state in the state bounds for which the policy outputs an action in the action bounds. For example, for the LunarLander problem, a query could be to verify if when the y-position of the lander is below some threshold value, i.e, when the lander is close to the ground, there exists a state such that the tested policy would output the action of pushing towards the ground: if the solver outputs ``SAT'', then there is a risk that the lander crashes. 

Designing interesting queries covering all risks is an open problem, hence to evaluate the verification times of our baseline policies, we generate 500 random queries per environment by sampling state and action subspaces uniformily. Out of those queries we only report the verification times of ``UNSAT'' queries since to verify that, e.g., the lander does not crash we want the queries mentioned above to be ``UNSAT''. We also only verify instances of ReLU MLPs using \citep{maraboupy} for this experiment as verifying decision trees requires a different software \citep{z3} for which verification times would not be comparable.
\begin{figure}[ht]
    \centering
    \includegraphics[width=1\linewidth]{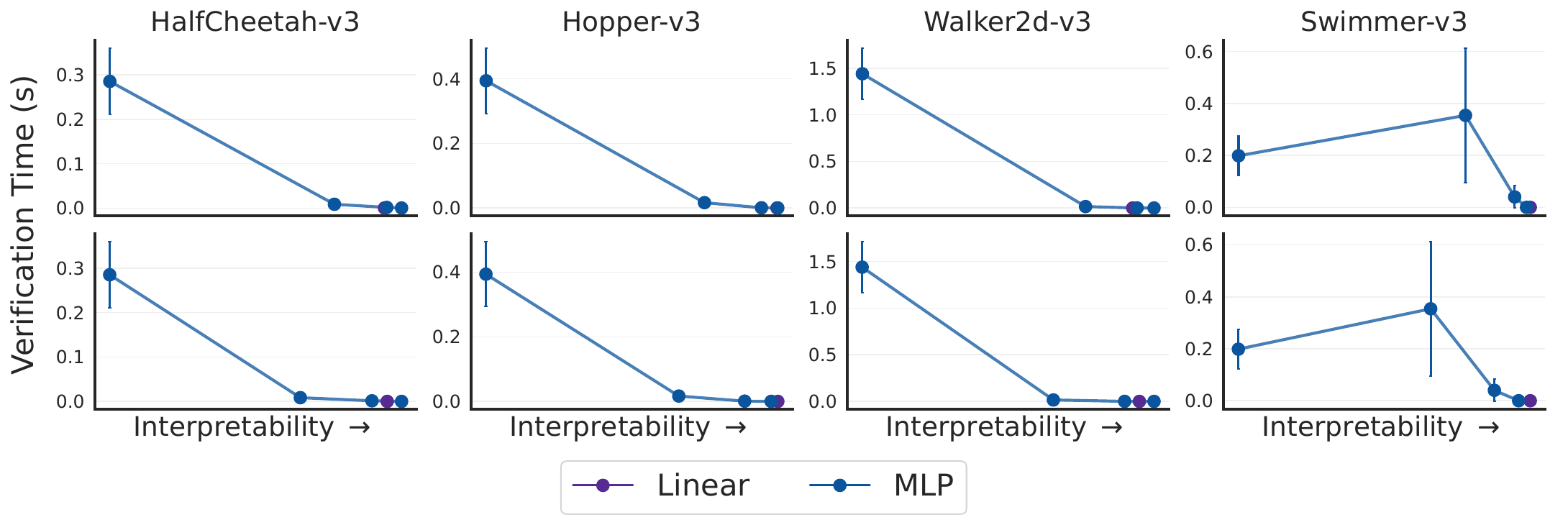}
    \caption{Verification time as a function of policy interpretability. Top row, interpretability is measured with step inference times. Bottom row, the interpretability is measured with policy size. We plot 95\% confidence intervals around means on both axes.}
    \label{fig:trade-off-verif}
\end{figure}

On Figure \ref{fig:trade-off-verif}, we can observe that our proxies for interpretability corroborates the claim of \citet{lens-complexity} that verification time decreases exponentially with MLP interpretability. This is another good validation of our proposed methodology as well as a motivation to learn interpretable policies. We discuss next some of our methodology's limitations.

\section{Limitations and conclusions}\label{sec:ccl}
We have shown that our proposed methodology provides researchers with a sound way of evaluating policy interpretability. In particular, we have shown that unfolding policies in a common language such as Python is a key component of our methodology to ensure that interpretability depends on the policy complexity (c.f. Figure \ref{fig:abl-proxies}). Furthermore, we were able to show that the proxies we use for interpretability leads to similar conclusions from user studies of interpretability or from other empirical evaluations of interpretability (c.f. Figures \ref{fig:abl-proxies}, \ref{fig:trade-off-summary}, and \ref{fig:trade-off-verif}). Using the proposed methodology, we were able to illustrate the trade-offs between episodic reward and interpretability of policies from different classes (c.f. Figure \ref{fig:trade-off-summary}) and showed the crucial need of our methodology as there is no better off policy class across tasks and metrics (c.f. Figures \ref{fig:perf-combined}, \ref{fig:abl-proxies}, and \ref{fig:trade-off-summary}). 

A nice property of our methodology is that it is independent of the learning algorithm of the interpretable policy. We chose imitation learning but it could have been a random search in the policies parameters space \citep{empirical-evidence}. Furhtermore, there sould be no limitation to use our methodology to evaluate the interpretability of arbitrary compositions of linear policies, trees and oblique trees, and MLPs, such as the hybrid policies from \citet{shindo2024blendrl}. However, the unfolded version of policies with loops which lengths depend on the state would change between step, hence, the policy size metric value will change during episodes. This is not necessarily a strong limitation but would require more work on the unfolding procedures as well as on defining episodic metrics. 

In the future, it would be interesting to compare episodic to averaged measures of interpretability. Indeed, we additionally show in Appendix \ref{fig:trade-off-episode} the interpretability-performance trade-offs using the inference time summed over entire episodes as the measure of interpretability. Even though using episodic inference does not change the trade-offs compared to step inference time, it is important to discuss this nuance in future work since a key difference between supervised learning and reinforcement learning interpretability could be that human operators would read policies multiple times until the end of a decision process. Using episodic metrics for interpretability is not as straightforward as someone would think as for some MDPs, e.g. Acrobot, the episodes lengths depend on the policy. We also did not evaluate the role of sparsity in the interpretability of linear and MLP policies even thought this could greatly influence the inference time. In the future it would be interesting to apply our methodologies to policies obtained with e.g. \citet{sparsity}. Moving away from evaluation, we also believe that our interpretable baselines can be used to train hierarchical agents \citep{hierarchical} using our baselines as options. We hope that our methodology as well as the provided baselines will pave the way to a more rigorous science of interpretable reinforcemeent learning.


\appendix
\section{Experimental details}

In this section we give all the experimental details necessary to reproduce our results.

\begin{table}[ht]
\centering
\small
\begin{tabular}{lll}
\hline
\textbf{Policy Class} & \textbf{Parameters} & \textbf{Training Algorithm} \\
\hline
Linear Policies & Determined by state-action dimensions & Linear/Logistic Regression \\
Decision Trees & [4, 8, 16, 64, 128] nodes & CART ($2\times$ nodes maximum leaves) \\
Oblique Decision Trees & [4, 8, 16, 64, 128] nodes & CART ($2\times$ nodes maximum leaves) \\
ReLU MLPs & [2$\times$2, 4$\times$4, 8$\times$8, 16$\times$16] weigths & Adam optimization (500 iterations) \\
\hline
\end{tabular}
\caption{Summary of baseline policy classes parameters and fitting algorithms (used in Line \ref{alg:fit}).}
\label{tab:policy-classes}
\end{table}

\begin{table}[ht]
\centering
\small
\begin{tabular}{lll}
\hline
\textbf{Classic} & \textbf{MuJoCo} & \textbf{OCAtari}\\
\hline
CartPole (4, 2, \textbf{490}) & Swimmer (8, 2, \textbf{300}) & Breakout (452, 4, \textbf{30})\\
LunarLander (8, 4, \textbf{200}) & Walker2d (17, 6, \textbf{2000}) & Pong (20, 6, \textbf{14})\\
LunarLanderContinuous (8, 2, \textbf{200}) & HalfCheetah (17, 6, \textbf{3000}) & SpaceInvaders (188, 6, \textbf{680})\\
BipedalWalker (24, 4, \textbf{250}) & Hopper (11, 3, \textbf{2000}) & Seaquest (180, 18, \textbf{2000})\\
MountainCar (2, 3, \textbf{90}) & \\
MountainCarContinuous (2, 1, \textbf{-110}) & \\
Acrobot (6, 3, \textbf{-100}) & \\
Pendulum (3, 1, \textbf{-400}) & \\
\hline
\end{tabular}
\caption{Summary of considered environments (dimensions of states and number or dimensions of actions, \textbf{reward thresholds}). The rewards thresholds are obtained from gymnasium \citep{gymnasium}. For OCAtari environments, we choose the thresholds as the minimum between the DQN expert from \citet{zoo} and the human scores. We also adapt subjectively some thresholds that we find too restrictive especially for MuJoCo (for example, the PPO expert from \citet{zoo} has 2200 reward on Hopper while the default threshold was 3800).}
\label{tab:envs}
\end{table}

\begin{table}
    \centering
    \footnotesize
    \begin{tabular}{c|cccccc}
    \toprule
    Envs & BC & BC & DAgger & DAgger & $Q$ & $Q$-DAgger\\
     & 50K & 100K & 50K & 100K & 50K & 100K\\
    \midrule
    Classic& 50 (PPO, DQN)& 50 (PPO, DQN)& 50 (PPO, DQN)& 50 (PPO, DQN)&  50 (DQN) & 50 (DQN)\\
    OCAtari& 0 & 0 & 0 & 5 (DQN)&  0 & 5 (DQN)\\
    Mujoco& 10 (SAC)& 10 (SAC)& 10 (SAC)& 10 (SAC)&  0 & 0\\
    \bottomrule
    \end{tabular}
    \caption{Repetitions of each imitation learning algorithm on each environment. We specify which deep reinforcement learning agent from the zoo~\citep{zoo} uses as experts in parentheses.}
    \label{tab:repet-distill}
\end{table}

\section{All interpretability-performance trade-offs}
In this appendix we provide the interpretability-performance trade-offs of all the tested environments. All the measures come from the experiment from Section \ref{sec:res-trade-offs}.

\begin{figure}
    \centering
    \includegraphics[width=1\linewidth]{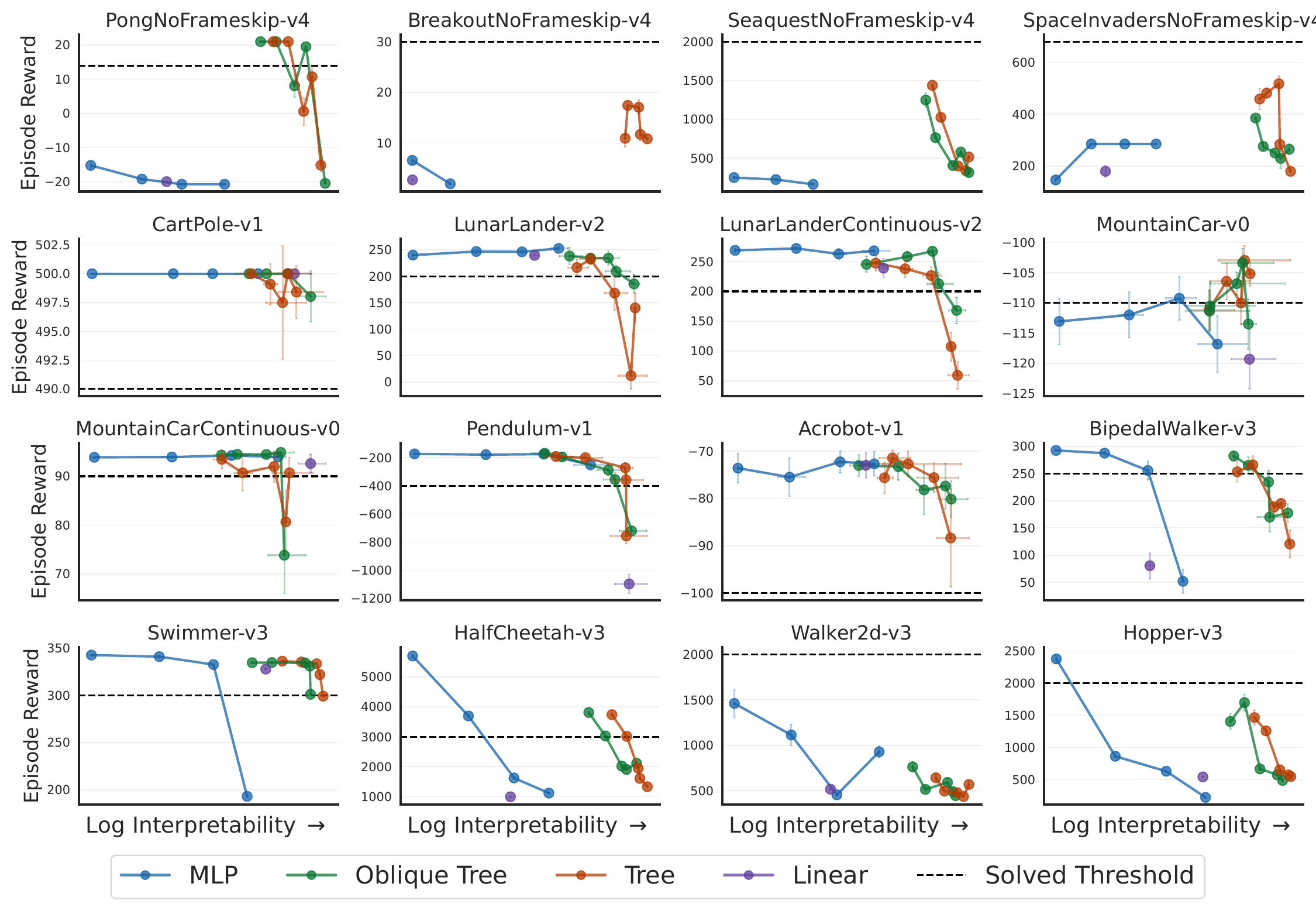}
    \caption{Trade-off Cumulative Reward vs. Step Inference Time}
    \label{fig:trade-off}
\end{figure}

\begin{figure}[ht]
    \centering
    \includegraphics[width=0.95\linewidth]{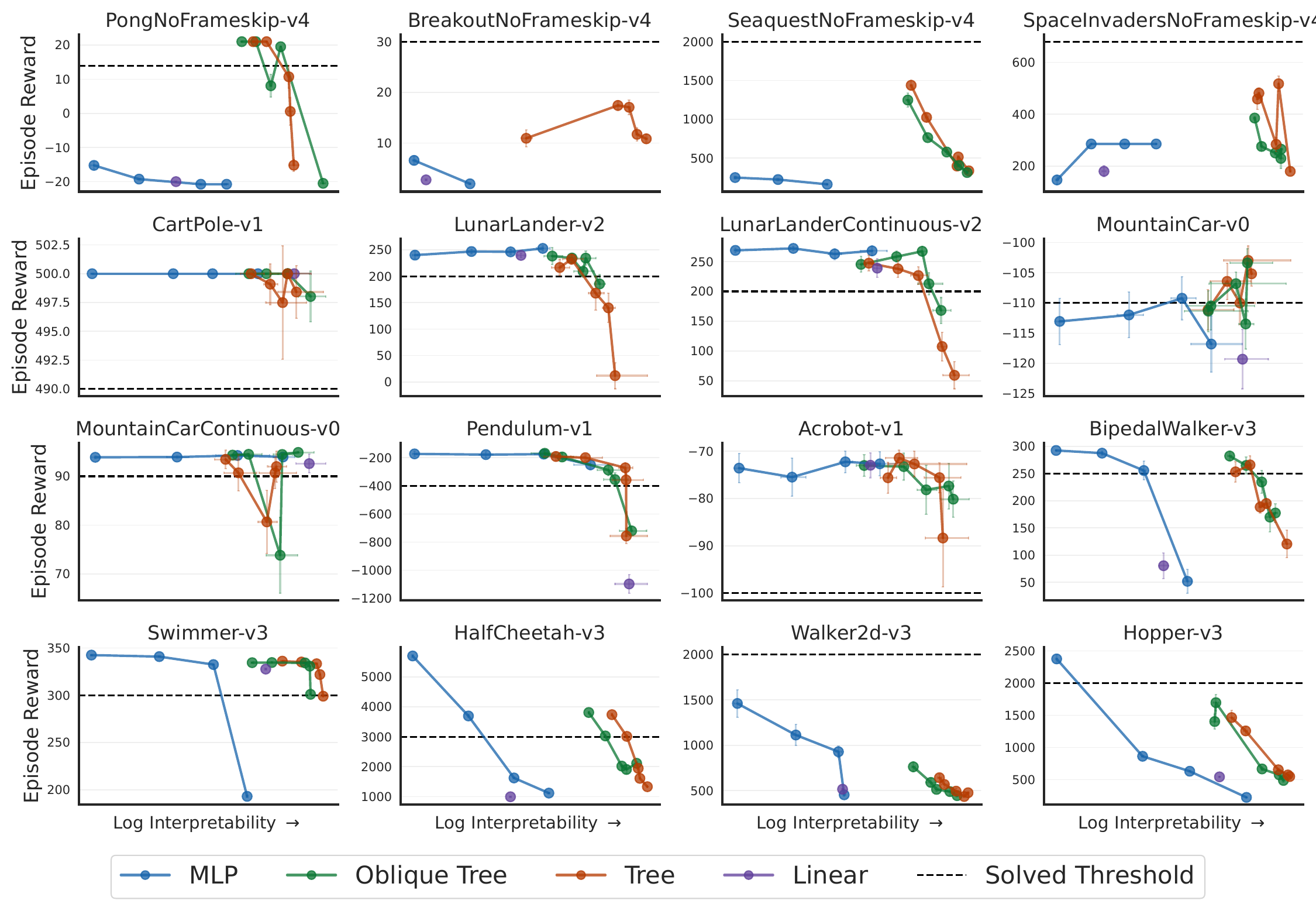}
    \caption{Trade-off Cumulative Reward vs. Episode Inference Time}
    \label{fig:trade-off-episode}
\end{figure}

\begin{figure}[ht]
    \centering
    \includegraphics[width=0.95\linewidth]{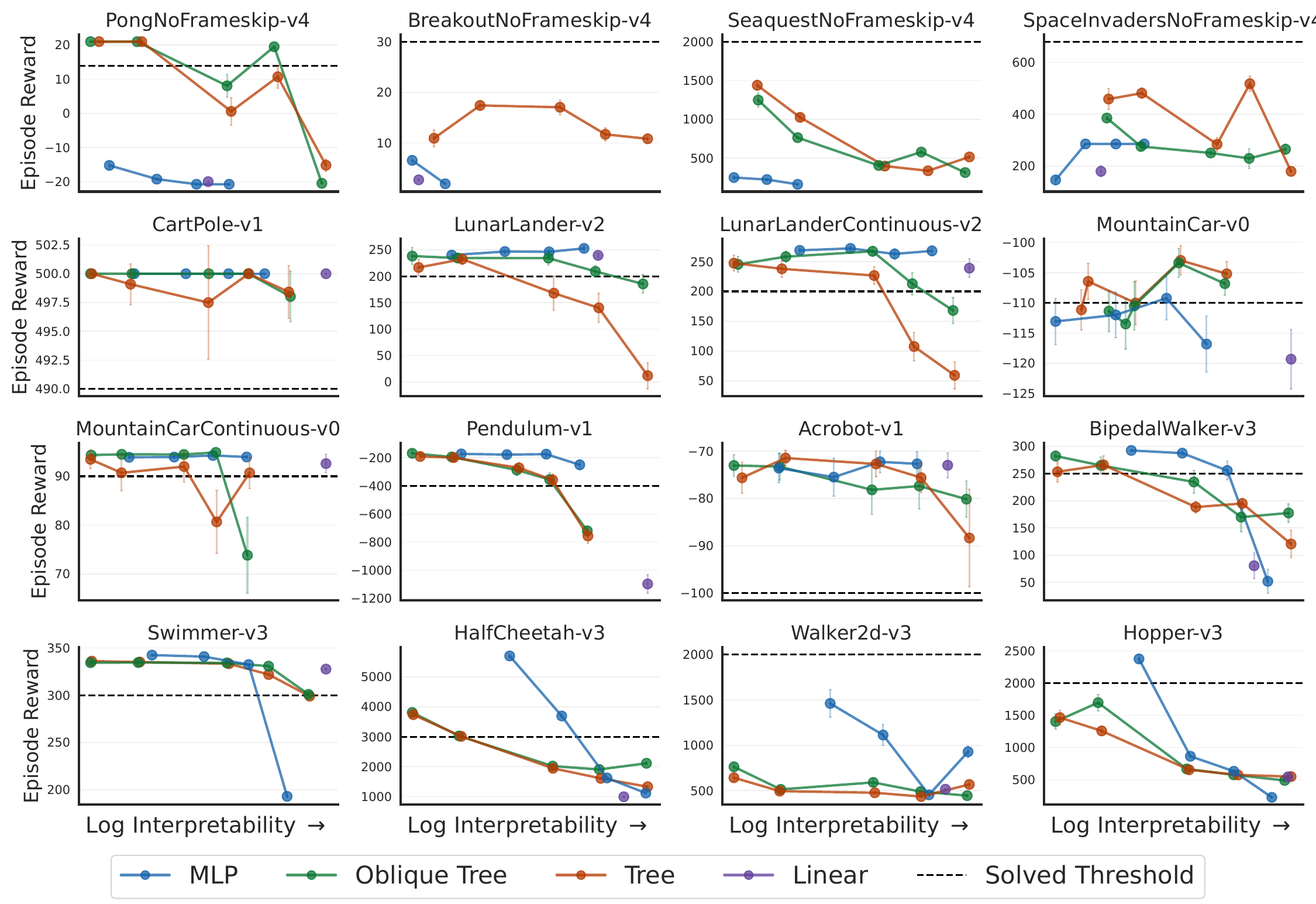}
    \caption{Trade-off Cumulative Reward vs. Policy Size}
    \label{fig:trade-off-size}
\end{figure}
\clearpage


\bibliography{main}
\bibliographystyle{rlj}


\end{document}